\documentclass{article}

\usepackage{microtype}
\usepackage{graphicx}
\usepackage{subcaption}
\usepackage{booktabs} 

\usepackage{hyperref}

\usepackage[accepted]{icml2026}

\usepackage{amsmath}
\usepackage{amssymb}
\usepackage{mathtools}
\usepackage{amsthm}

\usepackage[capitalize,noabbrev]{cleveref}

\theoremstyle{plain}

\theoremstyle{definition}

\theoremstyle{remark}

\icmltitlerunning{Dynamic Capability Scoping for Enterprise AI Agents}

\begin{document}

\twocolumn[
  \icmltitle{Dynamic Capability Scoping for Enterprise AI Agents: A Synthetic Dataset and Three-Source Permission Architecture}



  \icmlsetsymbol{equal}{*}

  \begin{icmlauthorlist}
    \icmlauthor{Halil Burak Noyan}{ind}
  \end{icmlauthorlist}

\icmlaffiliation{ind}{Independent Researcher, London, United Kingdom}

\icmlcorrespondingauthor{Halil Burak Noyan}{contact@buraknoyan.com}

  \icmlkeywords{AI Safety, Agentic Systems, Least Privilege, LLM Security, Synthetic Datasets, Dynamic Access Control, Indirect Prompt Injection, AI Alignment}

  \vskip 0.3in
]



\printAffiliationsAndNotice{}  

\begin{abstract}
Enterprise AI agents are typically granted static credential sets at configuration time, holding every tool the role might need for every task they perform. This persistent over-privilege expands the attack surface. We argue that capability scoping must follow a dynamic least-privilege principle and be treated as a prevention mechanism before a detection one. A credential that does not exist in an agent's context cannot be misused regardless of the agent's reasoning or evasion sophistication. We outline a three-source architecture instantiating this principle: role-based ceilings, a task-context classifier, and policy-derived combination prohibitions creating a layered proactive defense against LLM agent misalignment and misuse cases. The architecture supports both enforcing and observe-only deployment; the latter records agent permission requests inconsistent with task context, producing a behavioral signal usable in misalignment research.

As a first step toward evaluating this architecture, we 
contribute a synthetic dataset of 600 enterprise task prompts grounded 
in a multi-department company policy, labeled with minimum required 
permissions across a 15-permission tool-based taxonomy designed to map directly to 
deployable credentials or enforceable guardrails. The dataset is constructed via a two-pass pipeline that separates prompt generation from permission labeling to avoid circularity, and is validated against a 60-record/688 decisions human-reviewed sample (Cohen's $\kappa = 0.917$ pre-review and $\kappa = 0.967$ post-review). Iterating between dataset and policy reduced ceiling violations from 46 to 3, a 93\% reduction. This doesn't prove architectural performance but shows that synthetic prompt generation can drive policy refinement when the two are developed together. The dataset, environment specification, and generation pipeline are released to support evaluation of dynamic scoping mechanisms by the broader community.
\end{abstract}

\section{Introduction}

Enterprise deployments of AI agents typically grant a static capability envelope at configuration time. A fixed set of tools (file access, code execution, database queries, network calls, external email) are deployed at provision time that the agent holds for every task in its session, building on foundational architecture choices where models learn to invoke external APIs~\citep{schick2023toolformer, yaoetal2022react}. The envelope reflects what the role might ever need, not what any individual task actually requires. We call this gap \emph{context-privilege mismatch}. Preemptively deploying the permissions creates an unnecessary attack surface. A documentation-summarization task with no business reason to make a network call still has \texttt{http\_request} available, and a compromised or manipulated agent can use any tool that is already deployed on the agent environment or simply exfiltrate them. Recent work on alignment faking~\citep{greenblatt2024alignment} and frontier-model scheming~\citep{apollo2024scheming} shows that this is not a theoretical concern. Agents capable of reasoning about monitoring systems can adapt their behavior to evade detection, but they cannot use a credential that does not exist.

Existing approaches~\citep{inan2023llamaguard, rebedea2023nemo} to agent capability scoping treat the problem as one of input/output filtering or runtime monitoring. Policy-as-Prompt~\citep{kholkar2025policyasprompt} automates the translation of design documents into prompt-based classifiers that enforce least privilege at the agent boundary. Output filters and behavioral monitors observe agent actions and intervene when something looks wrong. Although detection mechanisms present a necessary layer of defense, they are reactive countermeasures instead of threat elimination at source. Against agents that reason about monitoring and adapt to evade it, detections have a structural ceiling because the attacker only needs to find one phrasing the monitor misses. Security risk and potential blast radius can be proactively minimized by implementing a dynamic least-privilege methodology. For example, if \texttt{http\_request} permission does not exist in the agent's context for the current task, a misaligned or compromised agent will lose the most straightforward capability to perform HTTP exfiltration. The attack surface is eliminated at the source, reducing the risk earlier in the cyber kill-chain than detective measures.

Effective capability scoping for AI agents requires multiple complementary layers, in the spirit of the Swiss cheese model of defense in depth~\citep{reason1990human}: a role-based ceiling that constrains an agent to the credentials its function legitimately requires, a task-context classifier that further narrows the permission set to what the current task actually needs, and a policy-derived prohibition layer that eliminates dangerous permission combinations regardless of any individual permission's legitimacy. The third layer specifically targets the \emph{lethal trifecta}~\citep{willison2025trifecta, greshake2023more} of private data access, exposure to untrusted content, and external communication, the conditions under which indirect prompt injection~\citep{greshake2023more} enables silent exfiltration. Each layer addresses a unique failure mode: role ceilings cannot distinguish between tasks, task classifiers can be manipulated by injection, and combination prohibitions are deterministic but coarse. Therefore they must be used in combination for effective attack surface minimization. The architecture supports both enforcing and observe-only deployment, described in Section~\ref{sec:combined}.

This paper presents the first of two contributions toward evaluating this architecture. We construct a synthetic dataset of 600 enterprise task prompts grounded in a multi-department company policy, labeled with minimum required permissions across a 15-permission taxonomy designed to map either directly to deployable credentials or agent guardrails. During validation, we observed that the two-pass pipeline inherently stress-tests the policy, exposing realistic workflow gaps that the policy fails to authorize. Iterating between the dataset and the policy reduced ceiling violations from 46 to 3 across the dataset, a 93\% reduction, demonstrating that synthetic prompt generation can drive policy refinement when the two are co-evolved. We release the dataset, the policy, and the validation results to support future evaluation of the full three-source architecture.

Our threat model consists of several LLM agent nodes deployed in hosts of an enterprise environment which has access to the inference model as well as various tools and capabilities to perform business-level operations on behalf of employees. For simplicity we are not including fully-autonomous service agents due to identity-without-role complications. The company has following departments: Engineering, Data and Analytics, Security, Customer Success, Finance, and Legal and Compliance. Expected workflow of each department is detailed in policy document both for policy-making and for automated enforcement. Departments have access to the following assets, which can contain 
business intellectual property, user data, employee information, and 
security-sensitive data: source code repositories (GitHub), issue 
tracking and project management (Jira), internal knowledge base 
(Confluence), customer relationship data (Salesforce), the analytics 
data warehouse (Snowflake, PostgreSQL), team communication channels (Slack), 
internal and external email, code execution environments, the public 
internet (HTTP), and uploaded session files. We are assuming tasks will have no gradual scope creep and discuss this limitation in the Section~\ref{sec:discussions}. Threat model assumes the task-context classifier, credential deployment service, and orchestrator are trusted components. Trusting an LLM-based classifier is a stronger assumption than trusting the deterministic orchestrator and credential service, and tightening it is left to future work.

Threat actors we are enumerating are misaligned LLM or a compromised agent. The attacker can attempt to invoke any tool the agent currently holds. The attack vectors within scope are indirect prompt injection, side-task execution, tool-output poisoning, malicious tool definition, LLM supply-chain attacks targeting the agent (as special case of misalignment). The mitigations against the misalignment and misuse are identical in our research context.

\section{Related Work}

Capability-based security is a methodology rooted in computer science 
since the 1960s~\citep{dennis1966programming}, with foundational work on least 
privilege~\citep{saltzer1975protection} and systems such as 
CapDesk~\citep{miller2006robust} and EROS~\citep{shapiro1999eros}. 
Classical capability-based systems trust the agent to hold its own 
credentials and attenuate them when delegating. In LLM agent security, 
the agent itself must not be trusted. We frame our research on how 
capability-based security applies to agentic workflows in AI control.

OWASP LLM Top 10 LLM06 (Excessive Agency)~\citep{owasp2025llmtop10} 
defines over-privileged agents as a fundamental problem in the AI safety 
domain. NIST SP 800-207~\citep{nist800207} specifies a zero trust 
architecture which treats each entity as untrusted by default regardless 
of perimeter. This forms the baseline of our research.

PCAS~\citep{palumbo2026pcas} proposes a policy compiler that provides 
an authorization layer based on a policy specification. This is the 
same idea as the policy enforcement component of our three-source 
permission scoping architecture detailed in Section 3. The significant 
difference is that whereas PCAS relies on the Datalog formal policy 
language, we argue LLM-generated rules compiled from official company 
policy are more performant and maintainable, in the spirit of 
Policy-as-Prompt~\citep{kholkar2025policyasprompt}.

\citet{zhang2026formalizing} outline a framework for the contextual 
security of LLM agents and point to the utility-security trade-off our 
overshoot/undershoot metrics measure directly. Our work contributes to 
that research direction by providing a realistic enterprise prompt 
dataset that can support evaluation of permission classifiers as one 
form of contextual defense.

\section{Three-Source Architecture}
\subsection{Role-Based Ceiling}
The first source is a deterministic lookup: given the agent's department, 
return the maximum permission set that role is authorized to hold. An 
Engineering agent may receive a database read credential because Engineering 
workflows include database queries; a Legal agent may not, because no Legal 
workflow in the company policy involves direct database access. The mapping 
is derived once from the organization's access policy. Source 1 is not a 
classifier and does no inference at runtime. Its function is to set a hard 
upper bound that no downstream source can exceed. Source 1 alone cannot 
distinguish between tasks within the same role: an Engineering agent 
summarizing a README and an Engineering agent debugging a production 
incident receive the same ceiling, even though their actual capability 
needs differ.

In multi-agent and autonomous deployments the agent is not necessarily 
tied to an employee role, and may have no department at all. We argue 
these cases should be modeled as service-level principals with their own 
IAM policies and their own ceilings, structurally identical to 
department-level ceilings but scoped to the agent's service identity 
rather than a human role. AWS implements this distinction directly through 
its IAM principal model~\citep{aws-iam-principals}: a principal is any 
entity (human or service) that can be authenticated and authorized, and 
ceilings attach to the principal regardless of its type. The Source 1 
mechanism is the same in both cases, only the principal that owns the 
ceiling differs.

\subsection{Task-Context Classifier}

The second source narrows the role-based ceiling to the permissions the 
current task actually requires. Given the user's task description and the 
agent's department context, a fine-tuned multi-label classifier predicts 
which of the 15 permissions in the taxonomy the task needs. Permissions 
outside that prediction are denied even when the role ceiling permits them. 
An Engineering agent asked to summarize a Confluence page receives 
\texttt{confluence\_read} and nothing else, despite the role ceiling 
allowing twelve other tools.

The classifier is a separate component from the agent itself. The agent 
does not declare its own required permissions or interact with the 
permissions classifier in any capacity. A compromised or deceptive agent 
would fabricate justifications to acquire maximum permissions. Source 2 
reasons over the task description independently and produces a prediction 
the agent cannot influence at runtime. This differs from prior context-aware approaches such as Conseca~\citep{tsai2025conseca}, which generates just-in-time policies per action from trusted context but evaluates the agent's own action attempts. Source 2 instead predicts the full task-required permission set up front from the user's task description.

The classifier outputs confidence 
scores per permission, and the orchestrator applies asymmetric thresholds 
based on permission risk: high-risk permissions such 
as \texttt{database\_read} or \texttt{email\_send\_external} are only granted 
at high confidence, while low-risk permissions such 
as \texttt{confluence\_read} are granted under lower confidence to avoid 
breaking common tasks. This dataset is the training input for that 
classifier; the architecture's empirical performance depends directly on 
the dataset's quality.

Source 2 has a known weakness due to operating on natural language. An 
attacker who can influence the task description through prompt injection, 
document pollution, or context manipulation can make the classifier grant 
higher permissions. Therefore we claim the task classifier is not a 
sufficient defense mechanism on its own, and must be complemented by role 
ceiling and policy based limitations. Source 2 is the most precise of the 
three mechanisms but also the most susceptible to attacks. A manipulated Source 2 can at worst grant permissions up to the role ceiling, excluding combinations prohibited by Source 3.

\subsection{Policy-Based Combination Prohibitions}
The third source is a deterministic filter applied to the output of 
Sources 1 and 2. It removes permission combinations that violate the 
organization's data handling policy. The motivating threat is the 
\emph{lethal trifecta} described by \citet{willison2025trifecta}: 
an agent simultaneously holding access to private data, exposure to 
untrusted content, and the ability to communicate externally is 
unconditionally exploitable via indirect prompt injection. Source 3 
makes the ``trifecta'' structurally impossible by prohibiting the 
co-occurrence of permissions that complete it. In TechCorp's policy, 
\texttt{database\_read} cannot co-occur with \texttt{email\_send\_external} 
or \texttt{http\_request}; when Sources 1 and 2 produce such a set, 
Source 3 removes the external-communication permission.

The prohibition rules can be authored manually from policy documents, but 
a more scalable approach uses Policy-as-Prompt~\citep{kholkar2025policyasprompt} to 
generate them automatically. A classifier reads the company policy and 
the expected workflow descriptions, then derives the set of permission 
combinations that fall within authorized workflows. Anything outside 
those combinations is either flagged or denied. This is a dynamic 
version of Source 3, the rules are not hand-written but inferred from 
the same policy document the rest of the architecture is grounded in. 
Updating the policy automatically updates the prohibitions.

Source 3 is a complementary mechanism that prevents inherently insecure 
workflows from being executed at the policy level. Consider a hypothetical 
vulnerability management agent with service-level privileges, tasked with 
a prompt such as ``search the internet for CVE-2026-1462 and check if 
our training pipeline uses the vulnerable Keras library.'' Such an agent 
may legitimately have access to model weights and the ability to make 
HTTP connections, providing a direct path for model weight exfiltration. 
A deceptive agent could exploit these permissions to execute a side task 
that bypasses Sources 1 and 2 entirely. Limiting internet access for 
agents that touch model weights is a policy-level decision that a 
policy-processing classifier can encode as a concrete prohibition rule.

\subsection{Combined Application of the Three-Source Architecture}
\label{sec:combined}
The three sources combine through intersection: a permission is granted 
only if Source 1 permits it (within the role ceiling), Source 2 predicts 
it (the task requires it), and Source 3 does not prohibit it (the 
combination is policy-compliant).

Each source's contribution can be measured independently by stepping 
through the layers. A baseline that grants the full department ceiling 
(Source 1 only) and a configuration that adds the task classifier 
(Sources 1 and 2) differ in exactly what Source 2 contributes; adding 
the policy filter on top isolates Source 3. We outline this evaluation 
as the empirical follow-up to this paper.

The architecture supports two deployment modes. In enforcing mode the 
orchestrator deploys only the predicted credentials and blocks tool calls 
outside that set. In observe-only mode it deploys the full role-based 
ceiling and lets the agent execute against it, but still computes the 
task-context and policy outputs and emits a structured event whenever 
the agent invokes a tool that the full system would have denied. 
Observe-only mode mirrors the permissive-mode pattern of access control 
systems such as SELinux~\citep{smalley2001selinux}, where policy is 
evaluated and logged before being enforced, and provides a deployment 
path that does not expose production workflows to undershoot. It also produces a behavioral signal that complements semantic monitors such as chain-of-thought review or output filtering. We see observe-only deployment as a route by which capability scoping contributes to alignment research, beyond its direct enforcement role.

\section{Synthetic Dataset Construction}
\subsection{Dataset Overview}
The dataset contains 600 enterprise task prompts grounded in a synthetic 
six-department company policy and labeled with the minimum permissions 
required from a 15-permission tool-based taxonomy. Prompts are generated 
through a two-pass pipeline that separates prompt generation from 
permission labeling to avoid circularity. A 60-record sample (10\%) was 
validated against expert review, achieving Cohen's $\kappa = 0.917$ pre-review and $\kappa = 0.967$ after disagreement resolution. The full dataset, the underlying company policy, the 
generation and labeling scripts, and the validation outputs are released 
at \url{https://github.com/0xballistics/mostargate} under GPL-3.0.

\subsection{The Company Policy}
The grounding policy describes a fictional mid-sized SaaS company, 
\emph{TechCorp}, with six departments: Engineering, Data and Analytics, 
Security, Customer Success, Finance, and Legal and Compliance. Each 
department's section in the policy specifies its tooling stack, typical 
workflows, and external communication patterns. A single-department policy would not 
exercise the full taxonomy, since no individual role needs every 
permission. Six departments produce natural coverage of the 15 
permissions through workflow diversity rather than synthetic profile 
construction. The complete policy text is included in the released 
repository.

\subsection{Two Pass Generation Pipeline}
The dataset is generated in two passes by separate LLM calls. Pass 1 
produces prompts only, given the company policy and a fixed department 
allocation, with no knowledge of the permission taxonomy. The Pass 1 
prompt also injects situational seeds (end-of-quarter crunch, production 
incidents, ad-hoc executive requests) alongside tone variations and 
occasional typos to drive prompt diversity, so the dataset reflects the 
texture of real enterprise messages with the diverse nature of the problems and 
communication styles. It includes both simple requests and complex 
multi-stage tasks that require reasoning and chained execution of tools.

Pass 2 takes each prompt and labels it with the minimum required 
permissions, given the policy, the prompt, and the full 15-permission 
taxonomy with labeling rules. Pass 2 labels permissions independently 
of the policy's combination prohibitions explicitly, as those are 
enforced by Source 3 at inference time, not encoded into the 
ground-truth labels.

This separation prevents spilling of prompt generation 
logic to labeling. A single pass approach would cause the model to 
produce prompts whose labels reflect how the prompt was phrased rather 
than what the task genuinely requires.

The Pass 1 department allocation is fixed at 5 Engineering, 4 Customer 
Success, 3 each for Data and Analytics, Security, and Finance, and 2 
Legal per batch of 20 prompts. Allocation is not proportional to 
headcount, smaller departments produce the highest-value high-sensitivity 
and external-communication tasks, so they are over-represented relative 
to size. Thirty batches yield 600 prompts. Generation prompts, labeling 
rules, and the full pipeline are released in the repository.

\subsection{Permission Taxonomy}
The 15 permissions can be grouped into five conceptual categories:

\begin{itemize}
  \item \textbf{Code:} \texttt{github\_read}, \texttt{pull\_request\_create}, \texttt{code\_execute}
  \item \textbf{Knowledge:} \texttt{confluence\_read}, \texttt{jira\_read}, \texttt{jira\_write}
  \item \textbf{Communication:} \texttt{slack\_read}, \texttt{slack\_write}, \texttt{email\_read}, \texttt{email\_send\_external}
  \item \textbf{Data:} \texttt{salesforce\_read}, \texttt{database\_read}, \texttt{http\_request}
  \item \textbf{Session:} \texttt{file\_read\_uploaded}, \texttt{export\_file}
\end{itemize}

These categories are descriptive, not structural --- the underlying taxonomy is a flat set of 15 tools.

The taxonomy follows two design rules. First, every permission corresponds 
to a deployable credential type (e.g.\ a Slack bot token, a GitHub PAT, a 
database access certificate) so the classifier output maps directly to 
what an orchestrator would actually issue at runtime. A grouped permission 
like \texttt{internal\_search} cannot be deployed as a single credential 
and was rejected for this reason. Second, read and write are split 
wherever the security implications differ: read operations primarily 
create exfiltration risk, while write operations create persistence and 
impersonation risk. The split is applied where the company policy 
distinguishes the two operations as separate workflows; for 
\texttt{database\_read} no corresponding write permission is included, 
because the TechCorp policy does not authorize agents to write to 
databases under any circumstances.

Each permission is assigned to one of three risk tiers, used by the 
classifier to apply asymmetric confidence thresholds at inference time: 
Tier 1 (default deny), Tier 2 (grant with justification), Tier 3 
(default permit). Tier assignments do not align with the conceptual 
categories above --- for example, \texttt{pull\_request\_create} is 
Tier 1 despite belonging to the Code category, and the two Session 
permissions split across Tier 2 (\texttt{export\_file}) and Tier 3 
(\texttt{file\_read\_uploaded}). The full taxonomy with credential 
types, risk tiers, and per-department ceilings is defined in 
\texttt{mostargate/constants.py} in the released repository.

\subsection{Adaptive Policy Refinement}
After the first complete generation run, validation against the 
department permission ceilings revealed 46 instances where a generated 
prompt required a permission absent from the requesting department's 
ceiling. These ceiling violations were not malformed prompts, in fact 
every flagged prompt described a plausible workflow for the department 
it was assigned to. The violations indicated that the ceilings, derived 
from a top-down reading of the company policy, were systematically 
under-specified relative to realistic enterprise workflows.

The two-pass pipeline made this mismatch visible. Pass 1 generates 
prompts without knowledge of the permission taxonomy, so it has no 
incentive to stay within ceiling boundaries. The prompts it produces 
reflect what employees in the policy's described departments would 
plausibly ask for, regardless of what the ceiling permits. When Pass 2's 
labels disagreed with the ceiling, the disagreement was diagnostic: 
either the prompt was unrealistic, or the ceiling was incomplete. We 
found the latter dominated.

Each violation was reviewed against the company policy and the relevant 
department's stated workflows. Five resolutions emerged: three ceiling 
expansions and two policy clarifications. The expansions added 
\texttt{email\_read} to Engineering, Security, and Data and Analytics 
(all three departments routinely receive external email despite the 
initial ceilings only granting \texttt{email\_read} to 
externally-corresponding roles); added \texttt{http\_request} to Data 
and Analytics (analysts fetch live external data to enrich reports); 
and added \texttt{jira\_read} and \texttt{jira\_write} to Finance 
(finance operations require issue tracking the initial policy did not 
describe). The two clarifications were the inverse case; workflows the 
model surfaced that should not be authorized. Salesforce CRM data is 
replicated to the Snowflake warehouse and should be queried via 
\texttt{database\_read} rather than \texttt{salesforce\_read}; Legal's 
GDPR data subject requests must be routed through Data and Analytics 
rather than direct database access. In both cases the policy was 
updated to describe the correct workflow, and the model was constrained 
at the source rather than by ceiling expansion.

After regeneration with the refined policy and ceilings, ceiling 
violations dropped from 46 to 3, a 93\% reduction. The remaining three 
were genuine edge cases that did not warrant either policy or ceiling 
change.

Synthetic prompt generation is not only 
a labeling-data source. When generation is decoupled from the policy 
that constrains the labels, the prompts function as a stress test of the 
policy itself, surfacing workflow gaps that top-down policy authoring 
routinely misses. The 93\% reduction shows that policy and dataset must 
be co-evolved: a fixed policy paired with a generated dataset will 
produce ceiling violations that corrupt the Source 1 signal, while 
iterating between them produces a policy aligned with how the 
organization actually operates and a dataset whose labels are 
internally consistent with that policy.

\section{Validation Results}
A 10\% random sample (60 records) was independently re-labeled by a 
single expert reviewer, with no access to the LLM's permission decisions 
for those records. Disagreements between the human and LLM labels were 
identified computationally afterwards. Each disagreement was then asked 
to the reviewer in a separate pass; the reviewer analyzed the LLM 
justification and tagged each as either \emph{llm\_correct} (human 
labeling error), \emph{human\_correct} (LLM prediction error), or 
\emph{ambiguous}. The corrected labels form the post-review ground truth.

Independent re-labeling prevents anchoring bias. A review-style 
validation, where the human inspects the LLM's labels and accepts or 
rejects them, anchors the human to the LLM's prior decisions. Our labeling process produces the human's own assessment of what the task requires, 
which the LLM's labels are then compared against. Later the disagreements are surfaced and resolved as 
a separate step. Both pre-review and post-review statistics are used to evaluate the results of the research.

Because most permissions are denied for most tasks (only 2-4 of the 15 are typically required), raw agreement shows inflated misleading accuracy. We used Cohen's $\kappa$ which is the standard reliability 
metric for two-rater categorical labeling and has been since Cohen 
introduced it in 1960~\citep{cohen1960coefficient}. We compute $\kappa$ over the in-ceiling decision space which means for 
each record, only permissions within the agent's department ceiling 
are evaluated. Otherwise out-of-ceiling permissions would be trivially 
denied by both raters and would inflate both observed and chance 
agreement. Adjusting department ceilings across 60 records yields 688 
binary human-LLM decisions. The reported $\kappa$ is computed once 
over those decisions; per-tool $\kappa$ values are also released for 
diagnostic use, but the headline figure is the overall $\kappa$. We 
additionally report the 95\% confidence interval using the standard 
error formula for Cohen's $\kappa$ from Fleiss, Cohen \& 
Everitt~\citep{fleiss1969kappa}.

Two metric sets are reported within the scope of research. \textbf{Pre-review} 
metrics treat the human labels as ground truth. \textbf{Post-review} 
metrics treat the reviewed labels as ground truth, which proved more accurate once the resolution data showed that human labels 
themselves contain errors. Both pre-review and post-review sets are released as part of this research.

Relying on a single labeler is a methodological limitation. Self-review bias is unavoidable here as the same 
researcher who produced the human labels also resolved their 
disagreements with the LLM. Cohen's $\kappa$ alone cannot rule out 
unconscious confirmation of prior decisions. We mitigate this by 
releasing the full disagreement log with per-case notes, so external 
reviewers can audit the resolution data directly. A second independent 
labeler would let us compute true inter-rater $\kappa$ and is noted 
as the most direct strengthening of the validation argument.

\subsection{Permission Metrics}
Table~\ref{tab:pre-post-metrics} reports the LLM's permission-labeling metrics 
under both ground-truth definitions.

\begin{table}[htbp]
\centering
\begin{tabular}{lrr}
\toprule
\textbf{Metric} & \textbf{Pre-review} & \textbf{Post-review} \\
\midrule
Exact match rate & 68.3\% & 85.0\% \\
Hamming accuracy & 97.1\% & 98.8\% \\
Macro F1 & 0.920 & 0.966 \\
Cohen's $\kappa$ & 0.917 & 0.967 \\
95\% CI on $\kappa$ & [0.882, 0.953] & [0.944, 0.990] \\
Overshoot rate & 1.3\% & 0.2\% \\
Undershoot rate & 8.1\% & 4.4\% \\
Weighted overshoot & 15.0 & 2.0 \\
$N$ (decisions evaluated) & 688 & 686 \\
\bottomrule
\end{tabular}
\caption{LLM labeling performance metrics on the 60-record validated 
sample. Pre-review treats the human labels as ground truth; 
post-review treats the adjudicated labels as ground truth, with the 
two ambiguous cases excluded from computation.}
\label{tab:pre-post-metrics}
\end{table}

The pre-review $\kappa$ of 0.917 (computed under the conservative assumption that human labels are always correct) places agreement firmly within the ``almost perfect'' agreement band of Landis and Koch~\citep{landis1977measurement}. Its lower CI bound of 0.882 demonstrates that even the worst-case interpretation of the validation data places agreement well above the 0.80 threshold commonly cited as a benchmark for substantial agreement in two-rater categorical 
labeling. Post-review $\kappa = 0.967$, with the 95\% confidence interval bounded below at 0.944, reflects the accuracy of the released labels after disagreement resolution.

\begin{figure}[htbp]
\centering
\includegraphics[width=\linewidth]{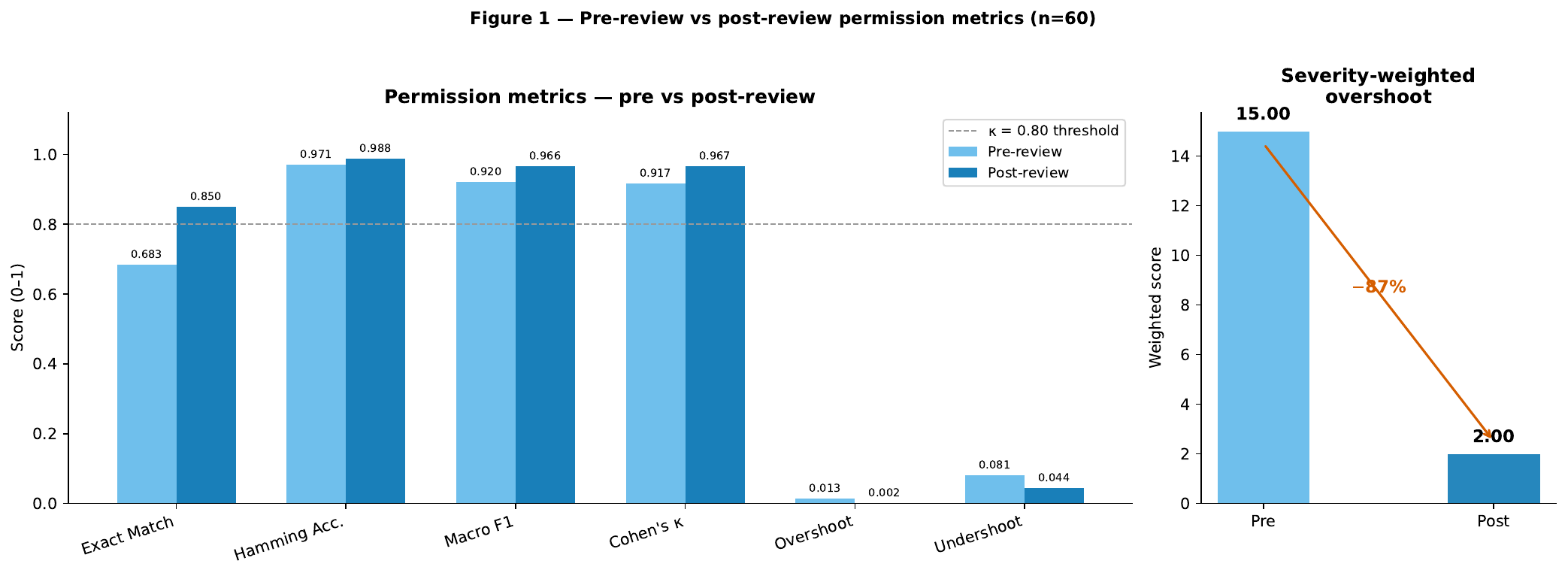}
\caption{Pre-review vs post-review permission metrics on the 60-record 
validated sample. Both $\kappa$ values sit in the ``almost perfect'' 
agreement band ($\kappa \geq 0.81$). The metric improvements after 
adjudication are modest in absolute terms ($\kappa$ moves 0.05), but 
the severity-weighted overshoot collapses from 15.0 to 2.0 (an 86.7\% 
reduction), which is the operationally consequential change.}
\label{fig:pre-post-metrics}
\end{figure}

The most security-relevant finding is the severity-weighted overshoot 
drop from 15.0 to 2.0, an 86.7\% reduction. Severity-weighted overshoot 
counts incorrectly granted permissions, weighted by their risk tier: 
a Tier 1 (default deny) over-grant counts three times a Tier 3 
over-grant. It essentially performs as a security risk score. The 
disagreement resolution attributed 10 of the 20 disagreements to human 
labeling error rather than LLM error, with the corrections concentrated 
on the high-risk permissions that drive the severity score. The 
post-review Figure~\ref{fig:pre-post-metrics} reflects the LLM's residual rate of high-risk 
over-granting on this sample.

The remaining error after resolution is conservative undershoot at 
4.4\%, corresponding to cases where the LLM denies a permission that 
should have been granted. For a permission classifier, undershoot is 
the preferred failure mode in principle: an unwarranted grant expands 
the attack surface, while a missed grant only degrades task performance. 
In real-world deployments undershoot disrupts workflow, so we expect 
production classifiers to be tuned with a directional bias matched to 
the deployment mode: undershoot bias under no enforcement, overshoot 
bias under enforcement, in both cases biased toward denying high-risk 
tools and permitting low-risk ones.

The two metric sets operate on slightly different decision spaces: 
pre-review evaluates 688 in-ceiling human--LLM decisions, while 
post-review evaluates 686 after excluding the two ambiguous cases that 
the disagreement resolution could not resolve. Both numbers exclude 
out-of-ceiling permissions, which are trivially denied by both raters 
and would inflate observed and chance agreement together.

Figure~\ref{fig:pre-post-metrics} visualizes the comparison.

\subsection{Disagreement Resolution}
20 of the 60 records contained at least one permission disagreement 
between the human and LLM labels. Each disagreement was reviewed in 
the resolution pass: 10 (50\%) were attributed to human labeling 
error, 8 (40\%) to LLM prediction error, and 2 (10\%) were judged 
ambiguous and excluded from post-review computation.

The near-parity between LLM-correct and human-correct cases is the 
methodological justification for using the reviewed labels as ground 
truth rather than the raw human labels. A clean oracle assumption 
would predict that the human ``wins'' disagreements most of the time; 
the actual 10/8 split shows that human labels contain errors at a 
rate comparable to LLM predictions on this sample.

The full disagreement log, including the resolution decision and a 
one-sentence rationale per case, is released with the dataset. 
External readers can audit the review directly.

\subsection{Sensitivity Tier Labels}
In addition to permissions, each generated record carries a sensitivity tier label - LOW, MEDIUM, or HIGH - assigned by the LLM at generation time based on the data being handled by the task. The tiers were originally intended to map to standard data-classification levels such as those in NIST SP 800-60~\citep{nist80060}, on the principle that more sensitive data implies higher security risk for any task that touches it. The validation pass found the label too subjective for that purpose: human-LLM raw agreement on sensitivity was 60\%, substantially below the corresponding figure for permission decisions on the same sample. The instability concentrates on the MEDIUM tier, which the confusion matrix shows is split nearly evenly across all three human assignments. The boundary between MEDIUM and the adjacent tiers is genuinely ambiguous in many enterprise tasks rather than a labeling defect. Of the 24 sensitivity disagreements adjudicated, the human was correct in 17 cases (71\%) and the LLM in 6 (25\%) with 1 ambiguous case (4\%), confirming that the LLM's sensitivity assignments are systematically less reliable than its permission assignments.

Sensitivity tier labels are therefore excluded from all classifier metrics and from the severity-weighted delta computation. They are retained in the dataset as descriptive metadata only.

\section{Discussion and Limitations}
\label{sec:discussions}
The release of this synthetic dataset and the methodology of adaptive policy refinement provide an empirical baseline toward evaluating dynamic capability scoping.

Our findings must be interpreted alongside several methodological considerations. First, our dataset validation currently relies on a single expert reviewer; while we released the full adjudication log to audit self-review bias, formal multi-rater validation is required to establish true inter-rater reliability. This will be handled in a future iteration. Second, despite careful grounding in a multi-department policy, synthetic data inherently lacks the full ecological validity, implicit context, and noise of real-world corporate communications. Third, because the task classifier relies on natural language processing, it inherits standard vulnerabilities to prompt injection. This reinforces our core design principle: dynamic scoping cannot function reliably in isolation and must be paired with deterministic ceilings and prohibitions. Finally, assigning data sensitivity tiers yielded lower human-to-LLM agreement (60\%), indicating that automating data classification requires more structured rubrics to be reliable.

A significant architectural limitation arises in the context of autonomous, multi-step agent workflows, such as ReAct~\citep{yaoetal2022react} or iterative tool-calling loops. Source 2, the task-context classifier, predicts required permissions based solely on the initial natural language prompt. However, complex open-ended tasks such as debugging a production incident often exhibit expanding permission scopes. An agent might initially only appear to need \texttt{jira\_read} to triage an alert, but upon reading the ticket, dynamically discover it requires \texttt{github\_read} to inspect a repository and \texttt{code\_execute} to reproduce a bug. Because Source 2 currently evaluates only the upfront request, it cannot anticipate downstream requirements discovered during the agent's reasoning loop. The orchestrator must either strictly undershoot (breaking the workflow) or preemptively over-provision tools for complex prompts, which risks increasing the dynamic scope back into the broad Source 1 role ceiling. We argue even open-ended tasks can still benefit from scope reduction in majority of cases (e.g., \texttt{slack\_write} permission within "debugging context" ).

The effectiveness of Source 3's prohibitions depends directly on the precision of the underlying rule mechanism and the workflow definitions in the organizational policy. Our baseline taxonomy blocks the co-occurrence of tools like \texttt{database\_read} and \texttt{http\_request} to prevent exfiltration. However, enterprise workflows frequently combine internal database reads with internal HTTP webhooks. If the rule mechanism lacks parameter-level scoping (e.g., distinguishing internal Virtual Private Cloud (VPC) endpoints from public IP addresses) or if the policy fails to explicitly distinguish authorized internal communication from prohibited external access, binary tool prohibitions will disrupt operations. Our adaptive policy refinement methodology directly addresses this requirement. Iterating between generated task prompts and policy definitions forces organizations to surface and resolve these network-scoping ambiguities prior to deployment, ensuring the architecture scales to complex enterprise environments.

Future work will advance this architecture from a design proposal to an empirically validated system. As an immediate next step, we will train a dedicated task-context classifier on this dataset to establish baseline overshoot and undershoot statistics. We will then evaluate the complete architecture across a phased progression (C0--C3): from an unenforced baseline (C0), to role ceilings only (C1), adding the task classifier (C2), and finally enforcing the policy prohibitions (C3). To further advance AI control against misalignment, we plan to deploy this setup as a custom micro-protocol within the ControlArena~\citep{control_arena} framework. Testing against deceptive model organisms~\citep{hubinger2024sleeper} will empirically evaluate the system's capacity to deterministically prevent hidden exfiltration side tasks. Lastly, we will explore practical operational integration via Just-in-Time credentialing and observe-only telemetry for enterprise security platforms.

\section{Conclusion}
Recent security incidents involving OpenClaw~\citep{openclaw2026systematic} demonstrate that agentic LLM deployments carry substantial security risks. As increasingly capable models are deployed within critical infrastructure, the impact of emerging misalignment grows exponentially. Anthropic reports that Claude Mythos~\citep{anthropic2026mythos}—their latest frontier model at the time of writing—demonstrates complex vulnerability exploitation and autonomous sandbox escape capabilities. Furthermore, sandbagging and alignment faking~\citep{greenblatt2024alignment} have the potential to bypass standard safety evaluations; a compromised agent that retains broad permissions can therefore cause severe damage. Substantial investment in AI control measures is required to mitigate these exploding risk factors.

Our proposal reduces the blast radius of these failure modes by removing the credentials a deceptive agent would require to execute an attack. A tool that does not exist in the agent's context cannot be misused, regardless of the agent's reasoning or strategic capabilities. This creates a partially-deterministic confinement layer that serves as a robust safety primitive, complementing existing alignment, monitoring, and detection solutions.

While the architecture introduces operational costs detailed in the Section~\ref{sec:discussions}, we expect these to be minimized through targeted fine-tuning. Despite current limitations in evaluating complex, open-ended tasks, we believe this architecture provides a critical component for the practical implementation of AI control measures in high-stakes environments.

\newif\ifaccepted
\acceptedfalse 

\ifaccepted
\section*{Acknowledgements}
The author would like to thank the organizers and community of the AI Security Bootcamp (AISB) Singapore for the technical discussions and threat modeling exercises that informed the Three-Source Architecture. Additional thanks go to BlueDot Impact for the research environment and mentorship provided during the project development. All views expressed are solely those of the author and do not represent the positions of his employer.
\fi

\bibliography{dynamic_scoping_icml}
\bibliographystyle{icml2026}

\end{document}